\documentclass{article}
\usepackage{spconf,amsmath,graphicx}

\usepackage{subfigure} 
\usepackage{amsmath,amsthm}
\usepackage{slashbox}
\usepackage{multirow}
\usepackage{amssymb}

\usepackage{hyperref}

\title{Joint Visual Denoising and Classification using Deep Learning}

\name{Gang Chen$^{\dagger}$ \qquad Yawei Li$^{\star}$ \qquad Sargur N. Srihari$^{\dagger}$}

\address{$^{\dagger}$ Department of Computer Science, SUNY at Buffalo, Buffalo NY 14260\\
    $^{\star}$ School of Communication and Information Engineering \\
    University of Electronic Science and Technology of China \\Chengdu, Sichuan 611731 China}
\begin{document}
%
\maketitle
\begin{abstract}
Visual restoration and recognition are traditionally addressed in pipeline fashion, i.e. denoising followed by classification. Instead, observing correlations between the two tasks, for example clearer image will lead to better categorization and vice visa, we propose a joint framework for visual restoration and recognition for handwritten images, inspired by advances in deep autoencoder and multi-modality learning. Our model is a 3-pathway deep architecture with a hidden-layer representation which is shared by multi-inputs and outputs, and each branch can be composed of a multi-layer deep model. 
Thus, visual restoration and classification can be unified using shared representation via non-linear mapping, and model parameters can be learnt via backpropagation. Using MNIST and USPS data corrupted with structured noise, the proposed framework performs at least 20\% better in classification than separate pipelines, as well as clearer recovered images. The noise model and the reproducible source code is available at {\url{https://github.com/ganggit/jointmodel}}.
\end{abstract}

\section{Introduction}
\begin{figure}[t!]
\centering
\includegraphics[trim = 20mm 30mm 30mm 30mm, clip, width=8.5cm]{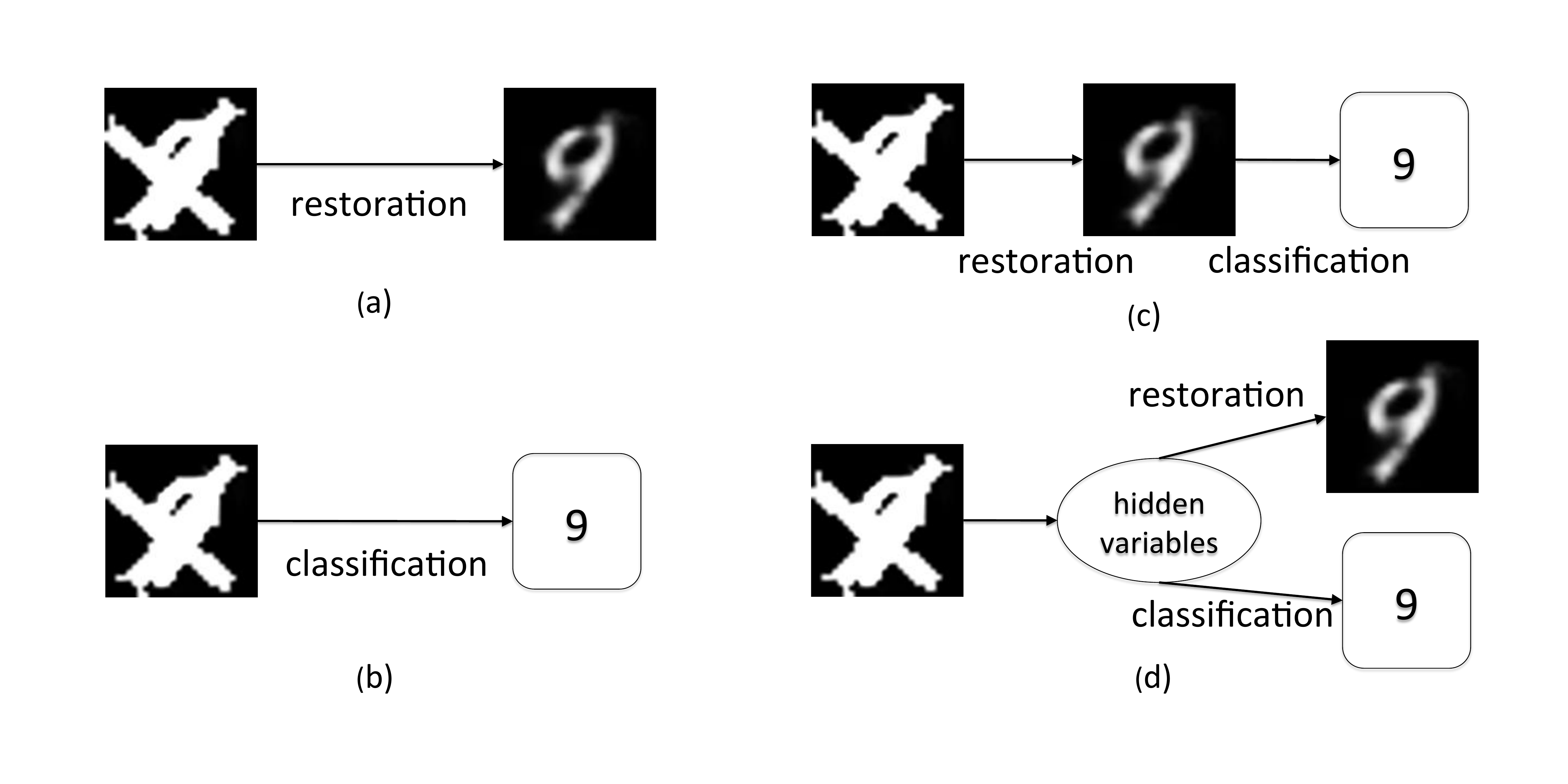}
\caption{Models for restoration/classification. (a) restoration; (b) classification; (c) pipeline; (d) joint restoration and recognition.}
\label{fig:pipeline}
\end{figure}

Common tasks in computer vision, such as image restoration and recognition, are usually regarded as separate tasks, shown in Fig. \ref{fig:pipeline}. In general, image restoration is an important problem whose purpose is to improve image quality in high-level vision tasks. And there is vast literature, most relying on unsupervised approaches, such as Wiener filter \cite{Wiener64}, Markov random field \cite{besag86}, sparse coding \cite{Elad06}, deep learning \cite{Xie12} with regularization terms or prior information of the underlying image. Visual recognition as a supervised task, has been extensively studied in machine learning and computer vision \cite{LeCun89,Larochelle12}. These two problems are characterized by very distinct statistical properties which make it difficult to address them together. Although they come from different input channels, there are connections between these two tasks: (1) the noisy image is derived from its clean one, (2) better image quality will improve recognition tasks. That is also the reason that we need preprocessing stage in many recognition problems. Hence, it is possible to learn useful representations which can potentially be used for such data to handle these two problems together. 

Recent advances in deep learning \cite{Hinton06b} and multi-modality learning \cite{Ngiam10} shed light on joint representation learning which captures the real-world concept that the data corresponds to. 
Deep learning \cite{Hinton06a,Bengio12} can learn abstract and expressive representations, which can capture a huge number of possible input configurations. 
The multimodal learning model \cite{Srivastava14b} in a sense extends the deep learning framework to handle different modalities. Thus it can learn a joint representation such that similarity in the code space indicates similarity of the corresponding concepts. 
However, these previous multi-modality models \cite{Ngiam10,Srivastava14b} can only handle one task. 
Moreover, how to jointly restore and classify images is also a challenge when the data is typically very noisy, e.g. structural noise.

We propose here a unified framework, which can learn a joint model to handle visual restoration and recognition together, refer to Fig. \ref{fig:pipeline}(d). Our one fan-in and two fan-out deep model is a network of 3 different kind of inputs coupled stochastic binary hidden units in a hierarchical structure. The inputs can be binary or real values, and they share a hidden layer via multi-layers non-linear mapping for each input. 
Specifically, visual restoration in our work is supervised, where the hidden and nonlinear structural information is learnt from data, and can handle more complex situations, such as structural denoising or super resolution. Furthermore, classification depends on the shared representation which is correlated to both clear and corrupted inputs. 
We pretrain the model with contrastive divergence, followed with gradient descent (L-BFGS) to update model parameters. We test our model on character denoising (to remove structured noises) and recognition tasks, and show the advantages over other separate baselines. 

\section{Related work}
There is little work, which models visual restoration and recognition in a joint framework with deep learning. However, there is much literature either addresses one or the other. The visual restoration problem, especially image denoising and super resolution, focuses on improve the image quality and numerous denoising methods have been proposed \cite{Wiener64,besag86,RudinO94,Tomasi98,Portilla03,Elad06,Jain08,ChenXC12}. 
Recent advances in deep learning \cite{Hinton06a,Hinton06b} has attracted great attention in machine learning community, and has been used for visual restoration. For example, 
deep neural networks have been used for denoising and inpainting \cite{Xie12} and yield promising results. The deep denoising autoencoder \cite{Gang14b} extends the work \cite{Hinton06b,Vincent10} for image denoising by minimizing the reconstruction loss to recover the original image. 
Deep learning has also been used for classification tasks, such as character recognition \cite{LeCun89,Hinton06b}, document classification \cite{Larochelle08} and image recognition \cite{Krizhevsky12}. The basic idea \cite{Bengio12} is to leverage deep neural networks, such as deep autoencoder, convolutional neural network \cite{LeCun89} or deep Boltzmann machines \cite{Salakhutdinov12} to learn representations helpful for classification. Recently, multi-modality learning, which generalizes deep learning to handle different input channels, has attracted great attention. For example, \cite{Srivastava14b} leverages deep Boltzmann machines to bridge images and texts. The model is a 2-fan (image and text as input or output) deep structure and learns a shared representation for the two modalities. Similarly, \cite{Ngiam10} et al. leverages deep autoencoder for multimodal deep learning, to handle video and audio data. 
Recently, a robust Boltzmann machine (RoBM) \cite{Tang12} was introduced for recognition and denoising. This model added another shape RBM to the Gaussian RBM prior to model the noisy variables which indicate where to ignore the occluder in the image. 
However, the experiments only show its effectiveness for regular structured noise. In this paper, we propose a unified framework, which can learn a joint model to handle visual restoration and recognition together. Our model is a 3-fan deep architecture, which generalizes previous multimodality models \cite{Ngiam10,Srivastava14b,ChenS15} for more complex multi-tasks, such as joint visual denoising and classification. 

\section{Joint model}
Our model for joint visual restoration and recognition is a 3-pathway deep architecture, with restricted Boltzmann machines (RBMs) as the building blocks. From another perspective, our model can be thought as the mixture of deep autoencoder and feedforward network. 
\subsection{Objective function}
In this part, we will present a jointly learning model for visual restoration and recognition. Assume that we have a training set $\mathcal{D} = \langle {\bf \tilde{v}}_{i},{\bf v}_i, {\bf y}_i \rangle_{i=1}^N$, with the corrupted image ${\bf \tilde{v}}_{i} \in \mathbb{R}_{+}^D$, the clear image ${\bf v}_{i} \in \mathbb{R}_{+}^D$ and its corresponding label ${\bf y}_{i} \in \{0,1\}^K$, for $i = \{1,2,...,N\}$. The purpose of our model is to learn a shared hidden representation in the deep architecture, which can restore the original image and label it given the noisy input. 
Thus, given the training triplet $\langle {\bf \tilde{v}}_{i}, {\bf v}_{i}, {\bf y}_{i} \rangle$, we use the following cross entropy loss:
\begin{align}
& \{{\theta_{i}\}_{i=1}^L, \{\theta^\prime_i\}_{i = 1}^L}, \{\theta^{\prime\prime}_i\}_{i = 1}^L  =\textrm{argmin}_{\theta, \theta^\prime, \theta^{\prime\prime}} \mathcal{L}({\bf \tilde{v}}_{i}, {\bf v}_{i}, {\bf y}_{i}; \theta, \theta^\prime, \theta^{\prime\prime})  \nonumber \\
&= \textrm{argmin}_{\theta, \theta^\prime, \theta^{\prime\prime}}   -  \sum_{i = 1}^N  {\bf v}_i \textrm{log} {\bf \hat{v}}_i +  (1- {\bf v}_i) \textrm{log} (1-{\bf \hat{v}}_i)  \nonumber \\
&  - \lambda \sum_{i = 1}^N  {\bf y}_i \textrm{log} {\bf \hat{y}}_i +  (1- {\bf y}_i) \textrm{log} (1-{\bf \hat{y}}_i) \label{eq:deepobj}
\end{align}
where $\{\theta, \theta^\prime, \theta^{\prime\prime} \}$ are the weights in the 3-way deep architecture respectively (we ignore the subscripts for clarity), $\lambda$ is the weight to balance the two losses. And ${\bf \hat{v}}_i$ and ${\bf \hat{y}}_i$ are the prediction from the noise input ${\bf \tilde{v}}_{i}$, specified as follows
\begin{align} 
& {{\bf h}_i} = \underbrace{f_{L} \circ f_{L-1} \circ \cdot\cdot\cdot  \circ f_1}_{L \textrm{ times}}({\bf \tilde{v}}_i) \label{eq:eqhidden} \\
 & {\bf \hat{v}}_i  =   \underbrace{g_1 \circ g_2\circ \cdot\cdot\cdot \circ g_{L}}_{L\textrm{ times}}  ({{\bf h}_i}) \label{eq:eqpred} \\
 & {\bf \hat{y}}_i  =   \underbrace{\phi_1 \circ \phi_2\circ \cdot\cdot\cdot \circ \phi_{L}}_{L\textrm{ times}}  ({{\bf h}_i}) \label{eq:eqlabels}
\end{align}
where $f_l$, $g_l$, and $\phi_l$ are non-linear projection functions, with weight parameters $\theta_l$, $\theta_l^\prime$ and $\theta_l^{\prime\prime}$ respectively in each layer. We ignore the underscript for parameters in mapping functions $f_l$, $g_l$, and $\phi_l$, for $l = \{1, ..., L\}$ in the above equations. As the deep belief network (DBN), we use the same logistic (or sigmoid) function in each layer. In our model, we attempt to learn the top layer hidden representations ${\bf h}_i$ via Eq. \ref{eq:eqhidden}, which are shared by the triplet $\langle {\bf \tilde{v}}_{i}, {\bf v}_{i}, {\bf y}_{i} \rangle$. In the predication stage, we hope to restore the clear image ${\bf v}_{i}$ and its label ${\bf y}_{i}$, using function compositions $\{g_l\}_{l=1}^L$ and $\{\phi_l\}_{l=1}^L$ respectively from the hidden layer ${\bf h}_i$.  Thus, we can think Eq. \ref{eq:eqhidden} as the encoding step, while Eqs. \ref{eq:eqpred} and \ref{eq:eqlabels} are the decoding steps. Our model has three pathways with the shared top layer $L$ in the center. 
In practice, we can select different number of layers and hidden nodes to predict ${\bf v}_{i}$ and ${\bf y}_{i}$ respectively. 

\subsection{ Learning and Inference}
Our deep model is a 3-branch deep structure with shared representations, which is different from the deep autoencoder. Thus, it is more complex to learn model parameters. In general, the parameters in each pathway can be pretrained separately in a completely unsupervised fashion, which allows us to leverage a large supply of unlabeled data. And the complexity of the pretraining depends on the number of layers and nodes in each pathway. 
For each pathway, we can initialize the weight parameters in each layer with DBN simultaneously. Then, we can infer the top hidden layer with mean field methods \cite{Salakhutdinov12} and update the top layer weights for each branch with constrative divergence (CD) \cite{Hinton06a}. After pretraining, we minimize the reconstruction error in Eq. (\ref{eq:deepobj}) with the global fine-tuning stage, which 
uses backpropagation through the whole network to compute gradients and then fine-tune the weights for optimal reconstruction and recognition (local minimum). Note that the gradients w.r.t. the shared hidden representations should be the summation from the two cross entropy losses in Eq. (\ref{eq:deepobj}). 

In the inference stage, for a corrupted image ${\bf \tilde{v}}$, we first use Eq. \ref{eq:eqhidden} to project it into the hidden space, and then reconstruct the clear ${\bf v}$ and predict its label ${\bf y}$.
\section{Experiments}
We analyzed our model on handwriting denoising problems on several standard handwriting datasets. We evaluated the denoising performance with Peak signal-to-noise ratio (PSNR) and recognition tasks with error rate.

\subsection{Data description}
{\bf The MNIST dataset}\footnote{\url{http://yann.lecun.com/exdb/mnist/}} consists of $28\times28$-size images of handwriting digits from $0$ through $9$ with a training set of 60,000 examples and a testing set of 10,000 examples, and has been widely used to test character denoising and recognition methods. A set of examples are shown in Fig. (\ref{fig:mnist_regular}). \\
{\bf The USPS Handwritten binary Alphadigits}\footnote{\url{http://www.cs.nyu.edu/~roweis/data/binaryalphadigs.mat}} are binary images with size $20 \times16$. There are digits of ``0" through ``9" and capital ``A" through ``Z", with 39 examples of each class. In our experiments, we only test our method on the binary Alphabets.

\subsection{Experimental setting}
In all experiments, we first use the stacked RBMs to initialize the model weights for all layers, and $\lambda=1$ to balance the two losses. In the fine-tuning state, we use the L-BFGS to optimize the model parameters. For the MNIST digits, we set the number of hidden nodes (encoding) [400 200 250 100] in the 4-layer deep model, the restoration output mirrors the setting except the last layer set as the same dimension as the input, and the recognition output has the same setting except the last layer set as the number of classes. For the USPS alphabets, we use the two layer deep encoding structure, with hidden nodes 100 and 64 respectively in each layer in the experiments. \\
{\bf Noise model}
We consider two kinds of structured noise that are widely appeared in the handwriting images. \\
{(1) The type 1 noise}: horizontal/vertical lines and sine waves, refer to Fig. \ref{fig:mnist_regular}(a) for visual understanding. \\
{(2) The type 2 noise}: random lines/strokes, refer the structural noise in Fig. \ref{fig:denoised}(b). 
Basically, the type 1 noise could corrupt images lightly, while the type 2 noise would heavily corrupt images, with more than 50\% regions. 
\\
{\bf Baselines}
We compare our method to Wiener \cite{Wiener64}, RoBM \cite{Tang12} and deep denoising autoencoder (DDAE) \cite{Gang14b}. 
\subsection{Results}
We first consider to remove the type 1 structural noise in the handwriting images. 
To generate the clean and noisy pairs, we add the type 1 noise to each MNIST image by randomly sampling horizontal/vertical lines or sine waves to construct its noisy observation. Then, we train our joint model on the 60,000 triplets (the clean, its noisy image and corresponding label), and test on the 10,000 noisy testing dataset for restoration and recognition. Analogously, we take the same way to train our model with type 2 noise.

\begin{figure}[t!]
\centering
\begin{tabular}{cc}
\includegraphics[trim = 48mm 99mm 45mm 95mm, clip, width=4.2cm]{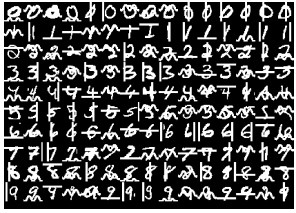} &
\includegraphics[trim = 48mm 99mm 45mm 95mm, clip, width=4.2cm]{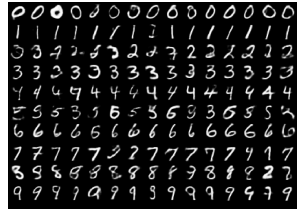} \\
(a) & (b) 
\end{tabular}
\caption{The denoising results comparison on the lightly corrupted MNIST dataset (the type 1 noise). (a) noisy images with horizontal/vertical lines/sine waves; (b) denoising results with our joint model. }
\label{fig:mnist_regular}
\end{figure}

\begin{figure}[h!]
\centering
\begin{tabular}{cc}
\includegraphics[trim = 48mm 99mm 45mm 95mm, clip, width=4.2cm]{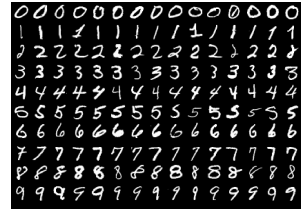}&
\includegraphics[trim = 48mm 99mm 45mm 95mm, clip, width=4.2cm]{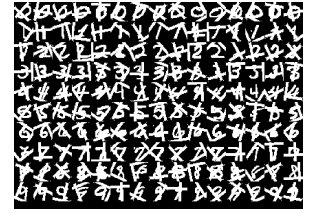} \\
(a) & (b) \\
\includegraphics[trim = 48mm 99mm 45mm 95mm, clip, width=4.2cm]{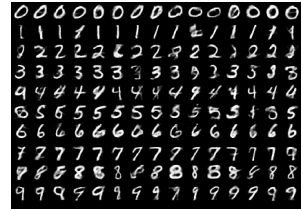} &
\includegraphics[trim = 48mm 99mm 45mm 95mm, clip, width=4.2cm]{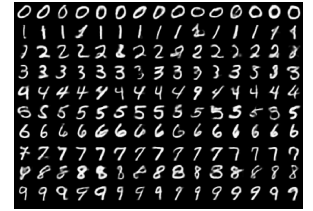} \\ 
(c) & (d)
\end{tabular}
\caption{The denoising results comparison on the heavily occluded MNIST digits (the type 2 noise). (a) original images; (b) noisy images with random structures; (c) denoising results with deep denoising autoencoder; (d) denoising results with our joint model. }
\label{fig:denoised}
\end{figure}

For the baselines, we first learn the deep neural network (DNN) \cite{Hinton06b} on the clean MNIST 60,000 images for classification, 
with default parameters, namely 4 layers with hidden nodes [500 500 2000 10] respectively for each layer. 
The error rate on the clean testing set we can get using DNN is 1.2\%, while the error rates on the noisy testing set are 41.9\% and 61.0\% respectively with the lightly and heavily corrupted noise in Figs. \ref{fig:mnist_regular}(a) and \ref{fig:denoised}(b). Then, we use the model learned to test the denoising baselines on the recognition task. 

The sampled results with our model are shown in Fig. \ref{fig:mnist_regular} (b), while the quantitative results were shown in Table (\ref{tab:mnistwave}). The lower bound of PSNR for the type 1 noise is $9.7$ dB, which is calculated on the noisy testing set. From the denoised results, we can see that our model is superior to the competitive baselines on both denoising and recognition. In other words, our joint model by leveraging label information for visual restoration is significant better than separate pipelines. We also test our method on the type 2 noise, refer noisy examples in Fig. \ref{fig:denoised}(b) and its denoised ones in Fig. \ref{fig:denoised}(d), as well as the quantitative performance in Table (\ref{tab:mnist}). 

\begin{figure}[h!]
\centering
\begin{tabular}{cccc}
\includegraphics[trim = 76mm 73.5mm 106.5mm 60mm, clip, width=1.8cm, height = 4.0cm]{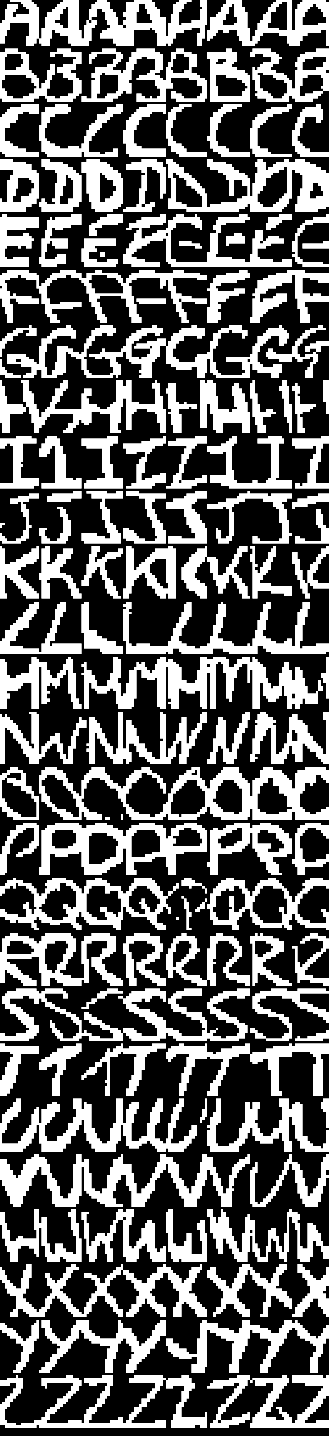}&
\includegraphics[trim = 76mm 73.5mm 106.5mm 60mm, clip, width=1.8cm, height = 4.0cm]{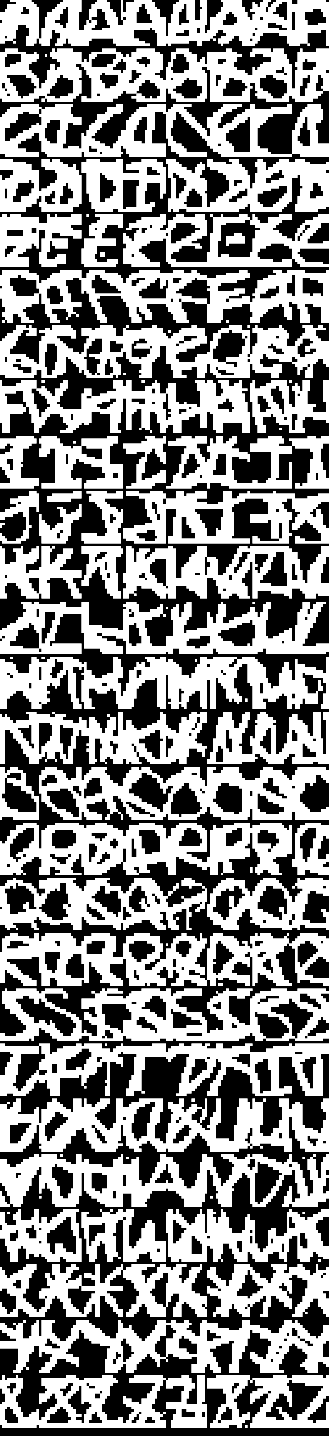} &
\includegraphics[trim = 76mm 73.5mm 106.5mm 60mm, clip, width=1.8cm, height = 4.0cm]{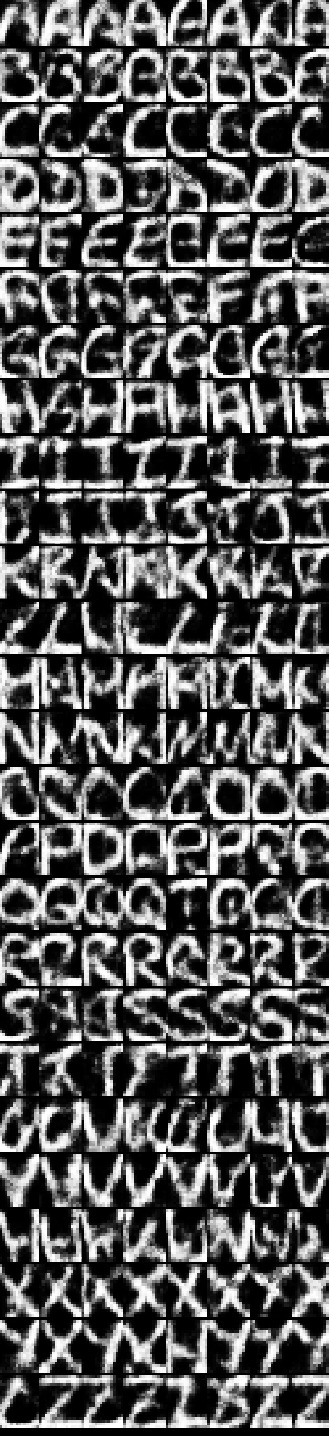} &
\includegraphics[trim = 76mm 73.5mm 106.5mm 60mm, clip, width=1.8cm, height = 4.0cm]{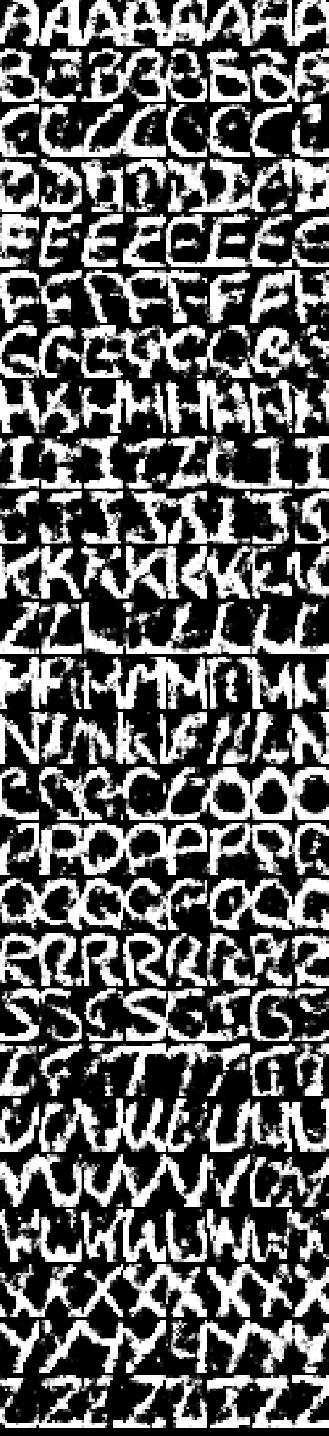} \\
(a) & (b) & (c) &(d)
\end{tabular}
\caption{The denoising results comparison on USPS alphabet (the type 2 noise). (a) original images from `A' to `Z' arranged in the top-down manner; (b) noisy images with random structures; (c) denoising results with deep denoising autoencoder; (d) denoising results with our joint model. }
\label{fig:usps}
\end{figure}

Apart from the digits, we also test our method on the USPS alphabets with the type 2 noise. Similar to the experiment on the MNIST digits, we add random strokes to the alphabets to create the noisy observations. Because there are only 39 training images for each class, we generated 10 corrupted samples for each clean image. Then we divided the clean and noisy binary pairs into the training set (account for 80\%) and testing set (the rest 20\%). We trained our joint model on the training set and test its performance on the testing set. The visual performance of our approach is shown in Fig. \ref{fig:usps} (d). The quantitative comparison between our method and the baselines is shown in Table (\ref{tab:usps}), which demonstrates that our method yields better denoising and labeling results. 

\begin{table}[t!]
\centering
\resizebox{.9\columnwidth}{!}{
\begin{tabular}{lrr}
\hline
Model & PSNR (dB) & Error rate (\%) \\
\hline
Wiener \cite{Wiener64} & 13.7 & 22.4\\
RoBM \cite{Tang12} & 15.8 & 17.2\\
DDAE w/o loop \cite{Gang14b} & 16.1 &  7.10 \\
DDAE with loop \cite{Gang14b}  & 17.7 &  4.36\\
Our method & 19.64 &  3.75 \\
\hline
DNN \cite{Hinton06b} & $\geq 9.7$ & $1.20 \sim 41.9$\\
\hline
\end{tabular}}
\caption{The experimental comparison on the lightly corrupted MNIST digits (type 1 noise). The PSNR value on the noisy testing set is 9.7 dB, which can be thought as the lower bound. The error rate range using DNN means that the error rates on the noisy testing set and the original clean testing set are 41.9\% and 1.2\% respectively. It demonstrates that our joint model can boost both denoising and recognition performance.}
\label{tab:mnistwave}
\end{table}

\begin{table}[t!]
\centering
\resizebox{.85\columnwidth}{!}{
\begin{tabular}{lrr}
\hline
Model & PSNR (dB) & Error rate (\%) \\
\hline
Wiener \cite{Wiener64} & 11.7 & 58.5\\
RoBM \cite{Tang12} & 13.9 & 52.6\\
DDAE w/o loop \cite{Gang14b} & 13.58 &  35.9 \\
DDAE with loop \cite{Gang14b} & 15.15 &  29.9\\
Our method & 18.6 &  12.7\\
\hline
DNN \cite{Hinton06b}& $\geq7.65$ & $1.20 \sim 61.0$\\
\hline
\end{tabular}}
\caption{The experimental comparison on the heavily corrupted MNIST digits (type 2 noise).} 
\label{tab:mnist}
\end{table}

\begin{table}[t!]
\centering
\resizebox{.85\columnwidth}{!}{
\begin{tabular}{lrr}
\hline
Model & PSNR (dB) & Error rate (\%) \\
\hline
Wiener \cite{Wiener64} & 14.2 & 67.8 \\
RoBM \cite{Tang12} & 16.3 & 62.8\\
DDAE w/o loop \cite{Gang14b} & 19.2&  42.5 \\
DDAE with loop \cite{Gang14b} & 18.5 &  44.1\\
Our method & 19.6 &  32.8\\
\hline
DNN \cite{Hinton06b}& $\geq 8.12$ & $1.29 \sim 67.4$\\
\hline 
\end{tabular}}
\caption{The experimental comparison on USPS alphabets (type 2 noise). The PSNR value of DNN is 8.12 dB, which shows the lower bound on the noisy testing set. The error rate range using DNN means that the error rates on the noisy testing set and the original clean testing set are 67.4\% and 1.29\% respectively.} 
\label{tab:usps}
\end{table}

\section{Conclusions}
In this paper, we consider the joint structural denoising and recognition problems on the handwriting images. We proposed a unified framework, which is a 3-fan deep architecture, and can learn the shared hidden representations for more complex multi-tasks. In a sense, our model can be thought as a mixture of deep autoencoder and deep feedforward neural network, which are unified in the joint framework for both reconstruction and classification tasks. The experimental results show the advantages of our model over competitive baselines on both denoising and recognition tasks.

\bibliographystyle{IEEEtran}
\bibliography{crbmbib}

\end{document}